\documentclass[runningheads,fleqn]{llncs}

\usepackage{latexsym}
\usepackage{amssymb}
\usepackage{amsmath}
\usepackage[all]{xy}
\usepackage{hyperref}

\newcommand{\ignore}[1]{}

\makeatletter
\newenvironment{prog}{\vspace{0.7ex}\par
\setlength{\parindent}{0.7cm}
\obeylines\@vobeyspaces\tt}{\vspace{0.7ex}\noindent
}
\makeatother
\newcommand{\startprog}{\begin{prog}}
\newcommand{\stopprog}{\end{prog}\noindent}

\newcommand{\id}{{\mathit{id}}} 

\newcommand{\ri}{{<\!\!\!<}}

\newcommand{\query}{\mathsf{query}}

\newcommand{\ground}{\mathit{G}}
\newcommand{\pfacts}{\mathit{prob\_facts}}

\def\defemb#1#2{\expandafter\def\csname #1\endcsname
                              {\relax\ifmmode #2\else\hbox{$#2$}\fi}}
\defemb{cA}{{\cal A}}
\defemb{cB}{{\cal B}}
\defemb{cC}{{\cal C}}
\defemb{cD}{{\cal D}}
\defemb{cE}{{\cal E}}
\defemb{cF}{{\cal F}}
\defemb{cG}{{\cal G}}
\defemb{cH}{{\cal H}}
\defemb{cI}{{\cal I}}
\defemb{cJ}{{\cal J}}
\defemb{cL}{{\cal L}}
\defemb{cM}{{\cal M}}
\defemb{cN}{{\cal N}}
\defemb{cO}{{\cal O}}
\defemb{cP}{{\cal P}}
\defemb{cQ}{{\cal Q}}
\defemb{cR}{{\cal R}}
\defemb{cS}{{\cal S}}
\defemb{cT}{{\cal T}}
\defemb{cU}{{\cal U}}
\defemb{cV}{{\cal V}}
\defemb{cX}{{\cal X}}
\defemb{cZ}{{\cal Z}}

\newcommand{\var}{\mathsf{var}}

\def\ri{<\!\!\!<}   

\def\res{\mathrel{\vert\grave{ }}}

\long\def\comment#1{}

\newcommand{\mgu}{\mathsf{mgu}}

\usepackage{xcolor}

\newcommand{\red}[1]{{\color{red} #1}}

\newcommand{\visible}{\mathit{visible}}
\newcommand{\xgen}{\mathsf{xgen}}

\begin{document}

\title{Explanations as Programs in\\ 
Probabilistic Logic Programming%
\thanks{This work has been partially supported by the EU (FEDER) 
and the Spanish MCI under grant 
PID2019-104735RB-C41/ AEI/10.13039/501100011033 (SAFER),
by the \emph{Generalitat Valenciana} under grant Prometeo/2019/098 
(DeepTrust), and by TAILOR, a project funded by EU Horizon 2020 
research and innovation programme under GA No 952215.
}
}

\author{Germ\'an Vidal}

\institute{
VRAIN, Universitat Polit\`ecnica de Val\`encia, Spain\\
  \email{gvidal@dsic.upv.es}
}

\titlerunning{Explanations as Programs in Probabilistic Logic Programming}

\maketitle

\pagestyle{headings}

\begin{abstract}
  The generation of comprehensible explanations is an
  essential feature of modern artificial intelligence systems.
  In this work, we consider \emph{probabilistic logic programming},
  an extension of logic programming 
  which can be useful to model domains with relational structure 
  and uncertainty. 
  Essentially, a program specifies a probability distribution 
  over possible \emph{worlds} (i.e., sets of facts). 
  The notion of \emph{explanation} 
  is typically associated with that of a world, so that one often 
  looks for the \emph{most probable world} as well as for the worlds 
  where the query is true.
  Unfortunately, such explanations exhibit no causal structure.
  In particular, the chain of inferences required for a specific 
  prediction 
  (represented by a query) is not shown. 
  In this paper, we propose a novel approach where explanations
  are represented as programs that are generated from a given query
  by a number of unfolding-like transformations. Here, the
  chain of inferences that proves a given query is made explicit.
  Furthermore, the generated explanations are minimal 
  (i.e., contain no irrelevant information) and 
  can be parameterized w.r.t.\
  a specification of \emph{visible} predicates, so that the user 
  may hide uninteresting details from explanations. \\[2ex]
  Published as \emph{Vidal, G. (2022). Explanations as Programs in Probabilistic Logic Programming. In: Hanus, M., Igarashi, A. (eds) Functional and Logic Programming. FLOPS 2022. Lecture Notes in Computer Science, vol 13215. Springer, Cham.}\\[2ex]  
  The final authenticated publication is available online at\\
   \url{https://doi.org/10.1007/978-3-030-99461-7\_12}
\end{abstract}

\section{Introduction} \label{sec:intro}

Artificial inteligence (AI) and, especially, machine learning systems 
are becoming ubiquitous in many areas, like medical 
diagnosis \cite{CG19}, intelligent transportation~\cite{VM20},
or different types of recommendation systems \cite{RRS15}, 
to name a few.
While prediction errors are sometimes acceptable, there are areas
where blindly following the assessment of an AI system
is not desirable (e.g., medical diagnosis). In these cases,
generating \emph{explanations} that are comprehensible
by non-expert users 
would allow them to verify the reliability of the prediction
as well as to improve the system when the prediction is not correct.
Furthermore, the last regulation on
data protection in the European Union \cite{GDPR16} has
introduced a ``right to explanation" for algorithmic decisions.
All in all, the generation of comprehensible explanations is an
essential feature of modern AI systems.

Currently, there exist  many approaches to 
\emph{explainable AI} (XAI) \cite{ARSBTBGGM20}, which
greatly differ depending on the considered application. 
In particular, so-called \emph{interpretable machine
learning} \cite{MCB20} puts the emphasis on the
interpretability of the models and their
predictions. 
In this work, we consider \emph{probabilistic logic programming} (PLP)
\cite{dRK15}, which can be useful to model domains with 
relational structure and uncertainty. 
PLP has been used for both inference---e.g., computing the marginal probability of a set of random variables given some evidence---and 
learning \cite{SK97,FBRSGTJR15}.
Among the different approaches to PLP, 
we consider those that are based
on Sato's \emph{distribution semantics} \cite{Sat95}.
This is the case of several proposals that combine logic
programming and probability, like  
Logic Programs with Annotated Disjunctions (LPADs)
\cite{VVB04}, ProbLog \cite{dRKT07}, 
Probabilistic Horn Abduction (PHA) \cite{Poo93},
Independent Choice Logic (ICL) \cite{Poo97},
and PRISM \cite{SK97}. 

In particular, we consider the ProbLog approach for its 
simplicity, but we note that the expressive power of all languages 
mentioned above is the same
(see, e.g., \cite[Chapter 2]{Rig18}).
A ProbLog program extends a logic program with a set
of \emph{probabilistic facts}. A
probabilistic fact has the form $p::a$ and denotes a
random variable which is true with probability $p$ and
false with probability $1-p$.  
Here, a program defines a probability
distribution over \emph{worlds}, i.e., sets of 
(possibly negated) atoms
corresponding to the probabilistic facts of the program. Essentially, 
the probability of a world is equal to the product of the 
probabilities of its true and false facts, while the
probability of a query is computed
by \emph{marginalization}, i.e.,  by summing up the probabilities
of the worlds where the query is true.

The notion of \emph{explanation} of a query 
is often associated with that
of a world. For instance, the MPE task \cite{SRVKMJ14},
which stands for \textit{Most Probable Explanation},  
consists in finding the world with the highest probability. 
However, a world exhibits no causal structure and, thus, 
it is up to the user to understand \emph{why} the given
collection of facts actually allow one to infer a particular
prediction (it might even be counterintuitive;
see Example~\ref{ex:win}).
Moreover, a world typically contains
facts whose truth value is irrelevant for the query,
which might be an additional source of confusion.
Alternatively, one could consider a \emph{proof tree} 
of a query as an explanation.
While the chain of inferences and the links to the
query are now explicit, proof trees are
typically very large and can be complex to understand by
non-experts. 

  In this paper, we propose a novel approach where explanations
  are represented as programs that are generated from a given query
  by a number of unfolding-like transformations. 
  In this way, we have the same advantages of using
  proof trees as explanations (the chain of inferences and the
  link to the query are explicit), but 
 they are often easier to understand by non-experts
  because of the following reasons: first, an explanation is
  associated with a \emph{single} proof, so it is conceptually
  simpler than a proof tree (that might comprise several proofs); 
  second, facts and rules have a more
  intuitive reading than a proof tree (and could easily be 
  represented using natural language); finally, the generated 
  explanations can be parameterized w.r.t.\ a set of \emph{visible}
  predicates. 
  If no predicate is visible, our explanations are not very different
  from a (partial) world, since they just contain the probabilistic
  facts that make a query true in a particular proof. 
  On the other hand, if all predicates are visible, the computation of 
  an explanation essentially boils down to computing the
  (relevant) grounding of a program for a given proof of the query.
  It is thus up to the user to determine the appropriate 
  level of detail in the explanation so that only the most
  interesting relations are shown.
  
  In this work, we 
  present a constructive algorithm for generating the explanations
  of a query such that the following essential properties hold: 
   i) the probability of each proof is preserved
  in the corresponding explanation, ii) explanations do not contain
  facts that are irrelevant for the given query,
  and iii) the (marginal) probability of the query in the original program is 
  equivalent to that in the union of the computed explanations.
  In order to check the viability of the approach, we have 
  developed a proof-of-concept implementation, 
  $\xgen$,\!\footnote{Publicly available from 
  \url{https://github.com/mistupv/xgen}\:.}
  that takes a ProbLog program
  and a query and produces a set of explanations  
  for this query, together with their associated probabilities. 
  
\section{Probabilistic Logic Programming (PLP)} \label{sec:plp}

In this section, we briefly introduce PLP following the ProbLog
approach (see, e.g., \cite{FBRSGTJR15,KDRCR11,dRK15,dRKT07,Rig18}
for a detailed account). 

Let us first recall some basic terminology from logic programming
\cite{Apt97,Llo87}. We consider a first-order language with a fixed
vocabulary of predicate symbols, constants, and variables
denoted by $\Pi$, $\cC$ and $\cV$, respectively.\footnote{We do not
consider function symbols in this work.}
An \emph{atom} has the form $f(t_1,\ldots,t_n)$ with $f/n \in \Pi$ 
and $t_i \in (\cC\cup\cV)$ for $i = 1,\ldots,n$.  
A \emph{definite clause} has the form $h\leftarrow B$, 
where $h$ (the \emph{head}) is an atom and $B$ (the \emph{body}) 
is a conjunction of atoms, typically denoted by a sequence
$a_1,\ldots,a_n$, $n\geq 0$;
if $n=0$, the clause is called a \emph{fact} and denoted 
by $h$; otherwise ($n>0$), it is called a \emph{rule}.
A \emph{query} is a clause with no head, and is denoted
by a sequence of atoms.
$\var(s)$ denotes the set of variables in the syntactic object $s$,
where $s$ can be an atom, a query or a clause.  A syntactic
object $s$ is \emph{ground} if $\var(s)=\emptyset$. 
Substitutions and their operations are defined as usual; 
the application of a substitution
$\theta$ to a syntactic object $s$ is denoted by
juxtaposition, i.e., we write $s\theta$ rather than $\theta(s)$. 

In the following, we consider that $\Pi = \Pi_p \uplus \Pi_d$, 
where $\Pi_p$ are the \emph{probabilistic predicates}
and $\Pi_d$ are the \emph{derived predicates}, which are disjoint. 
An atom $f(t_1,\ldots,t_n)$ is called a \emph{probabilistic atom}
if $f\in\Pi_p$ and a \emph{derived atom} if $f\in\Pi_d$.
A \emph{probabilistic logic program} (or just \emph{program} when
no confusion can arise)
$\cP = \cP_p\uplus\cP_d$ 
consists of a set of ground probabilistic facts $\cP_p$
and a set of definite clauses $\cP_d$.
A \emph{probabilistic fact} has
the form $p::a$, where $a$ is a ground atom and $p\in[0,1]$ 
is a probability such that
$a$ is true with probability
$p$ and false with probability $1-p$. 
These ground facts 
represent the Boolean random variables of the model, which
we assume mutually independent. 

In this paper, we also allow \emph{nonground} probabilistic
facts, that are replaced by their finite groundings using the
Herbrand universe, i.e., using the
constants from $\cC$. More generally, we 
consider \emph{intensional} probabilistic
facts defined by (probabilistic) clauses of the form
$p::f(x_1,\ldots,x_n) \leftarrow B$, where $B$ only contains
derived atoms. Such a rule represents
the set of ground probabilistic facts $p::f(x_1,\ldots,x_n)\theta$
such that $B\theta$ is true in the underlying 
model. 

\begin{example}  \label{ex:smokes}
Consider the following program 
(a variation of an example in \cite{FBRSGTJR15}):\footnote{We 
follow Prolog's notation in examples: variables start with an uppercase
letter and the implication ``$\leftarrow$" is denoted by ``\mbox{\tt :-}".} 
\begin{verbatim}
  0.8::stress(X) :- person(X).        person(ann). 
  0.3::influences(bob,carl).          person(bob).
  smokes(X) :- stress(X).
  smokes(X) :- influences(Y,X),smokes(Y).
\end{verbatim}
Here, we have two probabilistic predicates, 
\verb$stress/1$ and \verb$influences/2$, and a logic program
that defines the relation \verb$smokes/1$. Basically, the program
states that a person (either \verb$ann$ or \verb$bob$) is stressed 
with probability $\tt 0.8$, \verb$bob$ influences \verb$carl$ with
probability $\tt 0.3$, and that a person smokes either if (s)he 
is stressed or is influenced by someone who smokes.

Observe that the first probabilistic clause is equivalent to the following
set of ground probabilistic facts: \verb$0.8::stress(ann)$, 
\verb$0.8::stress(bob)$.
\end{example}
We note that probabilistic clauses can always be rewritten 
to a combination of probabilistic facts and non-probabilistic 
clauses \cite{FBRSGTJR15}. For instance, the probabilistic
clause in the example above could be replaced by
\begin{verbatim}
  0.8::p(X).       stress(X) :- person(X),p(X).
\end{verbatim}
In this work, we assume that the Herbrand universe is finite 
(and coincides with the domain $\cC$ of constants) and,
thus, the set of ground instances of each probabilistic fact is 
finite.\footnote{See Sato's seminal
paper \cite{Sat95} for the distribution semantics in the infinite case.}
Given a program $\cP$, we let $\ground(\cP)$ denote the set of its
\emph{ground} probabilistic facts (after grounding nonground
probabilistic and intensional facts, if any).
An \emph{atomic choice} determines whether 
a ground probabilistic fact is chosen or not. 
A \emph{total choice} makes a selection for \emph{all} 
ground probabilistic facts; it is typically represented as a
set of ground probabilistic atoms 
(the ones that are true). 
Note that, given $n$ ground probabilistic atoms, 
we have $2^n$ possible total choices. 

A program $\cP$ then defines a probability distribution over the
total choices.  Moreover, since the random
variables associated with the ground probabilistic facts
are mutually independent, the probability of a total choice
$L\subseteq \ground(\cP)$
can be obtained from the product of the probabilities
of its atomic choices:
\[
P(L) = \prod_{a\theta\in L} \pi(a) ~\cdot
\prod_{a\theta\in \ground(\cP)\setminus L} 1-\pi(a)
\]
where $\pi(a)$ denotes the probability of the
fact, i.e., $\pi(a)=p$ if $p::a\in\cP_p$.
A possible \emph{world} is then defined as the least Herbrand model
of $L\cup\cP_d$, which is unique. Typically, we denote a world by
a total choice, omitting the (uniquely determined) truth values for 
derived atoms.
By definition, the sum of the probabilities of all possible worlds
is equal to $1$.

In the following, we only consider \emph{atomic} queries. Nevertheless,
note that an arbitrary query $B$ could be encoded using an 
additional clause of the form $q\leftarrow B$. 
The probability of a query $q$ in a program $\cP$,
called the \emph{success probability} of $q$ in $\cP$, 
in symbols $P(q)$, is 
defined as the marginal of $P(L)$ w.r.t.\ query $q$:
\[
P(q) = \sum_{L\subseteq \ground(\cP)} P(q|L) \cdot 
P(L)
\]
where $P(q|L) = 1$ if there exists a substitution $\theta$
such that $L\cup \cP_d\models q\theta$ and $P(q|L) = 0$  otherwise.
Intuitively speaking, the success probability of a query is the
sum of the probabilities of all the worlds where this query
is provable.\footnote{Equivalently, has a successful SLD derivation;
see Section~\ref{sec:plpexplanations} for a precise definition
of SLD (\textit{Selective Linear Definite clause}) resolution.}

\begin{example} \label{ex:smokes2}
  \begin{figure}[t]
  \centering
  $
  \begin{array}{l|l@{}l@{}l@{}l|llll} 
  \multicolumn{1}{l}{~} &  & &&& P(w_i) \\\hline
  w_1 & \tt \{ stress(ann), & \tt stress(bob), & \tt influences(bob, carl) & \} & \tt 0.8\cdot 0.8\cdot 0.3 = 0.192\\
  w_2 & \tt \{ stress(ann), & \tt stress(bob) & \tt & \} & \tt 0.8\cdot 0.8\cdot 0.7 = 0.448\\
  w_3& \tt \{ stress(ann), & \tt & \tt influences(bob, carl) & \} & \tt 0.8\cdot 0.2\cdot 0.3=0.048\\
  w_4 & \tt \{ stress(ann) & \tt & \tt & \} & \tt 0.8\cdot 0.2\cdot 0.7= 0.112\\
  w_5 & \tt \{ & \tt stress(bob), & \tt influences(bob, carl) & \} & \tt 0.2\cdot 0.8\cdot 0.3 = 0.048 \\
  w_6 & \tt \{ & \tt stress(bob) & \tt & \} & \tt 0.2\cdot 0.8\cdot 0.7 = 0.112 \\
  w_7& \tt \{ & \tt & \tt influences(bob, carl) & \} & \tt 0.2\cdot 0.2\cdot 0.3 = 0.012 \\
  w_8 & \tt \{ \: & \tt & \tt & \} & \tt 0.2\cdot 0.2\cdot 0.7 = 0.028\\\hline
  \end{array}
  $
  \caption{Possible worlds for Example~\ref{ex:smokes}}   
  \label{fig:smokes}
  \end{figure}

  Consider again the program in Example~\ref{ex:smokes}.
  Here, we have eight possible worlds, which are shown in
  Figure~\ref{fig:smokes}. Observe that the sum of the 
  probabilities of all worlds is $1$.
Here, the query $\tt smokes(carl)$ is true
in worlds $w_1$ and $w_5$. Thus, its probability is 
$\tt 0.192+0.048= 0.24$.
\end{example}
Since the number of worlds is finite, one could compute the 
success probability of a query by enumerating all worlds and,
then, checking whether the query is true in each of them.
Unfortunately, this approach is generally unfeasible in practice
due to the large number of possible worlds. 
Instead, a combination of inference and 
a conversion to a Boolean formula is often used 
(see, e.g., \cite{FBRSGTJR15}).

\section{Explanations as Programs}

In this section, we focus on the notion of explanation in the
contect of PLP. Here, we advocate that a \emph{good} explanation 
should have the following properties:
\begin{itemize}
\item \emph{Causal structure}. An explanation should include
the chain of inferences that supports a given prediction.
It is not sufficient to just 
show a collection of facts. It should answer 
\emph{why} a given query is true, so that the user can
follow the reasoning from the query back 
to the considered probabilistic facts.

\item \emph{Minimality}. An explanation should not include irrelevant
information. In particular, those facts whose truth value is
indifferent for a given query should not be part
of the explanation.

\item \emph{Understandable}. The explanation should be easy to
follow by non-experts in PLP. Moreover, it is also desirable for
explanations to be parametric w.r.t.\ the information that is 
considered relevant by the user.
\end{itemize}
In the following, we briefly review some possible notions of an
explanation and, then, introduce our new proposal.

\subsection{Explanations in PLP} \label{sec:plpexplanations}

Typically, \emph{explanations} have been associated
with \emph{worlds}. For instance, 
the MPE (\textit{Most Probable Explanation})
task \cite{SRVKMJ14} consists in finding the world with the
highest probability given some \emph{evidence} (in our context, 
given that some query is true). 
However, a world does not show the chain of inferences of a given
query and, moreover, it 
is not minimal by definition, since it usually 
includes a (possibly large) number of probabilistic facts whose truth 
value is irrelevant for the query.

Alternatively, one can consider a probabilistic logic program
itself as an explanation. Here, the causal structure is explicit
(given by the program clauses). Moreover,
derived rules are easy to understand or can easily be explained
using natural language.
However,
the program explains the complete \emph{model} but it is
not so useful to explain a particular query: the chain of
inferences is not obvious and, moreover, programs are not usually
minimal since they often contain a (possibly large) 
number of facts and rules which
are not relevant for a particular query.

Another alternative consists in using 
the \emph{proof of a query} as an explanation. 
Following \cite{KDRCR11}, one can associate a \emph{proof} of a query
with a (minimal) \emph{partial} world $w'$ such that for all worlds
$w\supseteq w'$, the query is true in $w$. In this case, one can
easily ensure minimality (e.g., by using SLD resolution to determine
the ground probabilistic atoms that are needed to prove the query).
However, even if the partial world contains no irrelevant facts, 
it is still not useful to determine the chain of inferences behind
a given query. 
In order to avoid this shortcoming, one could represent the
proofs of a query by means of an SLD tree. Let us further
explore this possibility.

First, we recall some background from logic programming 
\cite{Llo87}. Given a logic program $\cP$, 
we say that $B_1,a,B_2 \leadsto_{\theta} (B_1,B,B_2)\theta$
is an \emph{SLD resolution step} if
$h \leftarrow B$ is a
renamed apart clause (i.e., with fresh variables) of program $\cP$, in
symbols, $h \leftarrow B \ri \cP$, and $\theta = \mgu(a,h)$
is the \emph{most general unifier} of atoms $a$ and $h$.
An \emph{SLD derivation} is a (finite or infinite)
sequence of SLD resolution steps. 
As is common, $\leadsto^\ast$ denotes the reflexive and 
transitive closure of $\leadsto$. In particular, we denote by
$A_0 \leadsto^\ast_\theta A_n$ a derivation 
$
A_0 \leadsto_{\theta_1} A_1 \leadsto_{\theta_2}
\ldots \leadsto_{\theta_n} A_n
$,
where $\theta =
\theta_1\ldots\theta_n$ if $n>0$ (and $\theta =\id$ otherwise).
An SLD derivation is called \emph{successful} if it ends with 
the query $\mathit{true}$ (an empty conjunction), and it is called
\emph{failed} if there is an atom that does not unify with the
head of any clause.
Given a successful SLD derivation
$A \leadsto^\ast_\theta \mathit{true}$, the associated \emph{computed
answer}, $\theta\!\res_{\var(A)}$, is the restriction of $\theta$
to the variables of the initial query $A$.
SLD derivations are represented by a (possibly infinite) finitely
branching tree called \emph{SLD tree}. 

All the previous notions (SLD step, derivation and tree, 
successful derivation, computed answer, etc)
can be naturally extended to deal with probabilistic logic programs
by simply ignoring the probabilities in probabilistic clauses.

Following \cite{KDRCR11}, the probability of a single proof
is the marginal over all programs where such a proof holds.
Thus, it can be obtained from the product of the
probabilities of the ground probabilistic facts used in the
corresponding SLD derivation.\footnote{Observe that  
each fact should only be considered once. E.g., given a successful
SLD derivation that uses the ground probabilistic fact
$\tt 0.4::person(ann)$ twice, the associated probability is $\tt 0.4$
rather than $\tt 0.4\cdot 0.4 = 0.16$.}
In principle, one could first apply a grounding stage---where 
all non-ground and intensional probabilistic facts
are replaced by ground probabilistic facts---and, then,
apply the above definition. 
Often, only a partial grounding is required 
(see, e.g., \cite{FBRSGTJR15}).
Since grounding is orthogonal to the topics of this paper,
in the following we assume that the following property holds:
for each considered successful SLD derivation 
$q\leadsto_\theta^\ast \mathit{true}$ that uses probabilistic 
clauses (i.e., probabilistic facts and rules) 
$p_1::c_1,\ldots,p_n::c_n$, we have that 
$c_1\theta,\ldots,c_n\theta$ are ground, i.e., it suffices 
if the probabilistic clauses used in the derivation \emph{become
eventually} ground. 

In practice, \emph{range-restrictedness} is often required for 
ensuring that all probabilistic facts become eventually ground
in an SLD derivation, 
where a program is range-restricted if all variables in the 
head of a clause also appear in some atom of the body
\cite{RS13}. 
Moreover, one can still allow some
probabilistic facts with non-ground
arguments (which are not range-restricted) as long as they 
are called with a ground term in these arguments;
see \cite[Theorem~1]{AR21}.
A similar condition is required in
ProbLog,
where a program containing a probabilistic fact of the
form \verb$0.6::p(X)$ is only acceptable if the query bounds
variable \verb$X$, e.g., \verb$p(a)$. However, if the
query is also non-ground, e.g., \verb$p(X)$, then
ProbLog outputs an error: ``Encountered a 
non-ground probabilistic clause".\footnote{The interested
reader can try the online ProbLog interpreter 
at \url{https://dtai.cs.kuleuven.be/problog/editor.html}.}

In the following, given a successful SLD derivation 
$D = (q\leadsto^\ast_\theta \mathit{true})$,
we let $\pfacts(D)$ be the set of ground probabilistic clauses 
used in $D$, i.e., $c\theta$ for each probabilistic clause $c$
used in $D$.
The probability of a proof (represented by a successful SLD derivation) 
can then be formalized as follows:

\begin{definition}[probability of a proof] \label{def:pd}
  Let $\cP$ be a  program and $D$ a 
  successful SLD derivation for some (atomic) query $q$ in $\cP$. 
  The probability of the proof represented by $D$ is obtained as follows:
  $
  P(D) = \Pi_{c\theta\in \pfacts(D)} ~\pi(c)
  $.
\end{definition}
Let us illustrate this definition with an example: 

\begin{example} \label{ex:smokes3}
  Consider again our running example. Here, we have the following
  successful SLD derivation $D$ for the query 
  $\tt smokes(carl)$:\footnote{We only show the relevant
  bindings of the computed $\mathsf{mgu}$'s in the examples.}
  \[
  \begin{array}{lll}
  \tt smokes(carl) 
  & \tt \leadsto & \tt influences(Y,carl),smokes(Y)\\
  & \leadsto_{\tt \{Y/bob\}} & \tt smokes(bob)\\
  & \leadsto & \tt stress(bob)\\
  & \leadsto & \tt \mathit{true}
  \end{array}
  \]
  Here, 
  $
  \pfacts(D) =\tt 
  \{  influence(bob,carl), ~stress(bob) \:\mbox{\tt :-}\: person(bob) \}
  $,
  whose probabilities are $\tt 0.3$ and $\tt 0.8$, 
  respectively. Hence, $P(D) = \tt 0.3\cdot 0.8 = 0.24$.
\end{example}
One might think that the probability of a query can then be computed
as the sum of the probabilities of its successful derivations. 
This is not generally true though, since the successful 
derivations may \emph{overlap} (e.g., two successful derivations
may use some common probabilistic facts).
Nevertheless, several techniques use the SLD tree as a first
stage to compute the success probability 
(see, e.g., \cite{KDRCR11,FBRSGTJR15}). 

Computing the most likely proof of a query  attracted
considerable interest in the PLP field
(where it is also called \emph{Viterbi proof} \cite{KDRCR11}).
Here, one aims at finding the most probable
partial world that entails the query (which can be obtained
from the proof with the highest probability). Note
that, although it may seem counterintuitive, the MPE cannot always
be obtained by extending the most likely proof of a query, as the
following example illustrates:

\begin{example} \label{ex:win}
Consider the following program from 
 \cite[Example 6]{SRVKMJ14}:
\begin{verbatim}
  0.4::red.    0.9::green.      win :- red, green.
  0.5::blue.   0.6::yellow.     win :- blue, yellow.
\end{verbatim}
Here, \verb$win$ has two proofs: one uses the probabilistic
facts \verb$red$ and \verb$green$, with probability 
$0.4\cdot 0.9= 0.36$, and another one uses the probabilistic
facts \verb$blue$ and \verb$yellow$, with probability
$0.5\cdot 0.6 = 0.30$. Hence, the most likely proof
is the first one, represented by the partial world
$\tt \{red,green\}$.
However, the MPE is the world
$\tt \{ green,blue,yellow\}$, with probability
$(1-0.4)\cdot 0.9\cdot 0.5\cdot 0.6 = 0.162$,
which does not extend the partial world $\tt \{red,green\}$.
This counterintuitive result can be seen as a drawback 
of representing explanations as worlds.
\end{example}
Considering proofs or SLD trees as explanations has obvious
advantages: they allows one to follow the chain of inferences
from the query back to the considered probabilistic facts and,
moreover, can be 
considered minimal. However, their main weaknesses 
are their complexity and size, which might be a problem 
for non-experts.

\subsection{Explanations as Programs} \label{sec:unfolding}

In this section, we propose to represent explanations
as programs. In principle, we consider that
rules and facts are easier to understand than proof trees for
non-experts (and could more easily be translated into natural 
language).\footnote{The use of rule-based models to explain the
predictions of AI systems is not new in the field of XAI
(see, e.g., \cite{ARSBTBGGM20}).}
Each program thus represents a minimal and more 
understandable explanation of a proof.
Moreover, the generation of explanations is now parametric w.r.t.\
a set of \emph{visible} predicates, thus
hiding unnecessary information.
We will then prove that the probability of a query in
an explanation is equivalent to that of the associated proof in
the original program, and that the marginal probability of a query is 
preserved when considering the union of all generated explanations.

The explanations of a query are essentially obtained using 
\emph{unfolding}, a well-known transformation in the
area of logic programming \cite{PP94}. Let 
$h\leftarrow B,a,B'$ be a clause and
$h_1\leftarrow B_1$,\ldots,$h_n\leftarrow B_n$ be \emph{all} the 
(renamed apart) 
clauses whose head unifies with $a$. Then, unfolding replaces
\[
h\leftarrow B,a,B'
\]
with the clauses
\[
(h\leftarrow B,B_1,B')\theta_1,~
\ldots,~
(h\leftarrow B,B_n,B')\theta_n
\]
where $\mgu(a,h_i)=\theta_i$, $i=1,\ldots,n$.
In the following, we assume that derived predicates are split into
\emph{visible} and \emph{hidden} predicates. In practice,
both predicates will be unfolded, but we introduce a special
treatment for visible atoms so that their calls are kept in the
unfolded clause, and a separate definition is added.
Intuitively speaking, visible predicates
represent information that the user considers relevant,
while hidden predicates represent intermediate or less relevant
relations that the user prefers not to see in an explanation. 

Given an atom $a$, we let $\visible(a)$ be true if $a$ is
rooted by a visible predicate and false otherwise.
The list of visible predicates 
should be given by the user, though a default specification
can be considered (e.g., our tool $\xgen$ assumes 
that all predicates are hidden unless otherwise specified).

The generation of explanations is modeled by a number
of transition rules. Given a query $q$, the initial explanation
has the form $\{\query(q) \leftarrow q\}$, where we assume
that $\query$ is a fresh predicate that does not appear 
in the considered probabilistic program. 
Then, we aim at unfolding this clause as much as possible.
However, there are some relevant differences with the 
standard unfolding transformation (as in, e.g., \cite{PP94}):
\begin{itemize}
\item First, we do not unfold the clauses of the 
original program but consider a new program
(i.e., the initial explanation). This is sensible in 
our context since we are only interested in those clauses
of the original program that are necessary for proving 
the query $q$. 

\item Second, we keep every nondeterministic unfolding 
separated in different explanations. This is due to the fact 
that our explanations represent a \emph{single} proof 
rather than a complete proof tree.

\item Finally, as mentioned above, we distinguish \emph{visible}
and \emph{hidden} predicates. While unfolding a hidden
predicate follows a standard unfolding, the case of visible
predicates is slightly different (see Example~\ref{ex:visible-unf} below).
\end{itemize}
During unfolding, we might find four different cases
depending on whether the considered clause is probabilistic
or not, and whether the considered atom is a derived or a
probabilistic atom. In the following, 
we consider each case separately. 

\subsubsection{Unfolding of derived atoms in derived clauses.}

This is the simplest case. Here, unfolding can be performed using
the following transition rules, depending on whether the 
atom is visible or not:
\[
\label{eqn:unfold1}
\begin{array}{c}
\displaystyle
\frac{\neg\visible(a)\wedge
h'\leftarrow B\ri\cP \wedge\mgu(a,h')=\theta
}{E\cup\{h\leftarrow B_1,a,B_2\}
\rightarrowtail (E\cup\{h\leftarrow B_1,B,B_2\})\theta } 
\tag{unf1}
\end{array}
\]
\[
\begin{array}{c}
\label{eqn:unfold2}
\displaystyle
\frac{\visible(a)\wedge
h'\leftarrow B\ri\cP \wedge\mgu(a,h')=\theta 
\wedge \rho(a\theta)=a'}{E\cup\{h\leftarrow B_1,a,B_2\}
\rightarrowtail E\theta\cup\{a'\leftarrow B\theta,
h\theta\leftarrow B_1\theta,\underline{a'},B_2\theta\}} 
\tag{unf2}
\end{array}
\]
where atoms marked with an underscore 
(e.g., atom $\underline{a'}$ in rule \ref{eqn:unfold2} above)
cannot be selected for unfolding anymore,
and $\rho$ is a simple renaming function that takes an atom 
and returns a new atom with a fresh predicate name 
and the same arguments
(e.g., by adding a suffix to the original
predicate name in order to keep its original meaning). For instance,
$\tt \rho(smokes(carl)) = smokes_1(carl)$. 
While rule (\ref{eqn:unfold1}) denotes a standard unfolding
rule (\ref{eqn:unfold2}) is a bit more involved. 
The fact that an atom is visible does not
mean that the atom should not be unfolded. It only means that
the call should be kept in the unfolded clause in order to
preserve the visible components of the inference chain.
Indeed, observe that the computed $\mgu$ is applied to \emph{all} 
clauses in the current explanation. This is sensible since 
all clauses in an explanation actually represent one single 
proof (i.e., a successful SLD derivation).

\begin{example} \label{ex:visible-unf}
  Consider the following logic program:
\begin{verbatim}
  p(X) :- r(X,Y).       r(X,Y) :- s(Y).       s(b).
\end{verbatim}
and the query $\tt p(a)$. 
A successful SLD derivation for this query 
is as follows:
\[
\tt p(a) \leadsto r(a,Y) \leadsto s(Y) \leadsto_{\{Y\mapsto b\}} true
\]
Given the initial explanation 
$E_0 = \tt \{\query(p(a)) ~\mbox{\tt :-}~ p(a)\}$,
and assuming that no predicate is visible, 
the (repeated) unfolding of $E_0$ using rule 
(\ref{eqn:unfold1}) would eventually produce the 
explanation $E' = \tt \{\query(p(a)) \}$,
which can be read as ``the query \verb$q(a)$ is true''.
In contrast, if we consider that $\tt r/2$ is visible,
we get the following unfolding sequence:
\[
\begin{array}{l}
E_0 = \tt \{\query(p(a)) ~\mbox{\tt :-}~ p(a)\}\\
E_1 = \tt \{\query(p(a)) ~\mbox{\tt :-}~ r(a,Y) \}
~\red{\mbox{\it //using rule $\tt p(X)~\mbox{\tt :-}~r(X,Y)$}}\\
E_2 = \tt \{r_1(a,Y) ~\mbox{\tt :-}~s(Y),
~\query(p(a)) ~\mbox{\tt :-}~ \underline{r_1(a,Y)} \} 
~\red{\mbox{\it //using rule $\tt r(X,Y)~\mbox{\tt :-}~s(Y)$}}\\
E_3 = \tt \{r_1(a,b),
~\query(p(a)) ~\mbox{\tt :-}~ \underline{r_1(a,b)} \}
~\red{\mbox{\it //using fact $\tt s(b)$}}\\
\end{array}
\]
The generated explanation ($E_3$) is a bit more informative
than $E'$ above: 
``the query $\tt p(a)$ is true because $\tt r(a,b)$ is true",
where $\tt r/2$ is renamed as $\tt r_1/2$ in $E_3$.
\end{example}
In the example above, renaming $\tt r/2$ is not really
needed. However, in general, 
the renaming of visible atoms is necessary to
avoid confusion when unfolding nondeterministic predicates,
since we want each explanation to represent one, \emph{and only one}, proof.
Consider, e.g., the following program:
\begin{verbatim}
  p :- q,q.    q :- a.   q :- b.   a.    b.
\end{verbatim}
ant the query \verb$q$.
Assume that predicate \verb$q/0$ 
is visible
and that the first call to \verb$q$ is unfolded using
clause \verb$q :- a$ and the second one using
clause \verb$q :- b$. Without predicate renaming, 
we will 
produce  an explanation including a clause of the form 
$\tt query(p) ~\mbox{:-}~ \underline{q},\underline{q}$,
together with the two clauses defining \verb$q$.
Unfortunately, this explanation does not represent a single
proof (as intended) since every call to \verb$q$ 
could be unfolded with either clause.
Renaming is then needed to ensure that only one
unfolding is possible:
$\tt query(p) ~\mbox{:-}~ \underline{q_1},\underline{q_2}$,
together with the 
clauses $\tt q_1 ~\mbox{:-}~ a$ and $\tt q_2 ~\mbox{:-}~ b$. 

\subsubsection{Unfolding of derived atoms in probabilistic clauses.}

As mentioned before, probabilistic rules are used 
to provide an intensional representation of a set of ground
probabilistic facts.
One could  think that the unfolding a derived
atom in such a clause will always preserve 
the probability of a query.
However, some caution is required:

\begin{example}
Consider the following program:
\begin{verbatim}
  q(a).   q(a).   0.8::p(X) :- q(X).
\end{verbatim}
By unfolding clause \verb$0.8::p(X) :- q(X)$, we would get the
following program:
\begin{verbatim}
  q(a).   q(a).   0.8::p(a).   0.8::p(a).
\end{verbatim}
Here, $P({\tt p(a)})=\tt 0.8$
in the first program but $P({\tt p(a)})=
\tt 0.8\cdot 0.2+0.2\cdot 0.8 + 0.8\cdot 0.8 = 
0.96$ in the second one.
\end{example}
The problem with the above example is related to the interpretation
of intensional facts. Observe that $\pfacts$ in Definition~\ref{def:pd}
returns a \emph{set}. This is 
essential to compute the right probability and to avoid 
duplicates when there are several successful derivations 
computing the same ground answer.

Therefore, if we want to preserve the  probability of a 
query, we can only unfold derived atoms when they do not
have several proofs computing the same ground answer.
Since we are assuming that derivations are finite, this property could
be dynamically checked. 
In the following, for simplicity, we assume instead that programs 
cannot contain several occurrences of the same (ground) 
probabilistic fact.\footnote{Nevertheless, our tool
$\xgen$ considers more general programs by requiring the
specification of those predicates that may violate the above
condition (see Section~\ref{sec:xgen}).}
The new unfolding rule is thus as follows:
\[
\label{eqn:unfold3}
\begin{array}{c}
\displaystyle
\frac{
h'\leftarrow B\ri\cP \wedge\mgu(a,h')=\theta
}{E\cup\{p::h\leftarrow B_1,a,B_2\}
\rightarrowtail (E\cup\{p::h\leftarrow B_1,B,B_2\})\theta } 
\tag{unf3}
\end{array}
\]

\subsubsection{Unfolding of a probabilistic atom in a
derived clause.}

In this case, one might be tempted to define the
unfolding of clause $h\leftarrow B_1,a,B_2$ using clause
$p::h'\leftarrow B$ and $\mgu(a,h)=\theta$ as the clause
$p::(h\leftarrow B_1,B,B_2)\theta$. However, such
a transformation would generally change the success probability
of a query, as illustrated in the following example:

\begin{example}
Consider the following program
\begin{verbatim}
       p :- a,b.     p :- b,c.     0.6::a.    0.7::b.    0.8::c.
\end{verbatim}
where \verb$p$ holds either because \verb$a$ and \verb$b$
are true or because \verb$b$ and \verb$c$ are true. By unfolding
\verb$b$ in the first clause of \verb$p$ using the strategy
above, we would get
\begin{verbatim}
  0.7::p :- a.       p :- b,c.     0.6::a.    0.7::b.    0.8::c.
\end{verbatim}
However, $P({\tt p})= 
P({\tt a,b}) + P({\tt b,c}) - P({\tt a,b,c}) =
\tt 0.6\cdot 0.7 + 0.7\cdot 0.8 - 0.6\cdot 0.7\cdot 0.8
= 0.644$ in the original program but
$P({\tt p}) = 0.7448$ in the unfolded one. 
Intuitively speaking, the issue 
is that, by embedding the probability
of \verb$b$ into the unfolded clause of \verb$p$, 
the worlds associated with the two proofs of \verb$p$
no longer overlap.
To be precise, the clause \verb$0.7::p :- a.$
is equivalent to \verb$p :- a,pp$ with \verb$0.7::pp$.
Thus, we now have 
$P({\tt p}) = 
P({\tt a,pp}) + P({\tt b,c}) - P({\tt a,pp,b,c}) = \tt 
0.6\cdot 0.7 + 0.7\cdot 0.8 - 0.6\cdot 0.7\cdot 0.7\cdot 0.8 
= 0.7448$. 
\end{example}
Therefore, in the following, probabilistic atoms are 
always (implicitly) considered as \emph{visible} atoms: 
\[
\begin{array}{c}
\label{eqn:unfold4}
\displaystyle
\frac{p::h'\leftarrow B\ri\cP \wedge\mgu(a,h')=\theta 
}{E\cup\{h\leftarrow B_1,a,B_2\}
\rightarrowtail (E \cup\{p::h'\leftarrow B,
h \leftarrow B_1,\underline{a},B_2\})\theta} 
\tag{unf4}
\end{array}
\]
Note that the probabilistic atom is not renamed, in contrast
to the renaming of visible atoms in rule (\ref{eqn:unfold2}) above.
Renaming would be required only if a program could
have several probabilistic atoms with different probabilities,
thus introducing some undesired nondeterminism
(but we ruled out this possibility, as mentioned before).

\subsubsection{Unfolding of a probabilistic atom
in a probabilistic rule.} 

Although this situation cannot happen
in the original program (we required intensional facts to have only
derived atoms in their bodies), such a situation may show up after
a number of unfolding steps. This case is similar to unfolding a
probabilistic atom in a derived clause:
\[
\begin{array}{c}
\label{eqn:unfold5}
\displaystyle
\frac{p'::h'\leftarrow B\ri\cP \wedge\mgu(a,h')=\theta 
}{E\cup\{p::h\leftarrow B_1,a,B_2\}
\rightarrowtail (E \cup\{p'::h' \leftarrow B,
p::h \leftarrow B_1,\underline{a},B_2\})\theta} 
\tag{unf5}
\end{array}
\]
In the following, given some initial explanation $E_0$, we refer
to a sequence $E_0 \rightarrowtail E_1 \rightarrowtail \ldots \rightarrowtail
E_n$, $n\geq 0$, as an \emph{unfolding sequence}.
If further unfolding steps are possible, we say that 
$E_n$ is a \emph{partial explanation}. Otherwise, 
if $E_n \not\rightarrowtail$, we have two possibilities:
\begin{itemize}
\item If the clauses in $E_n$ contain no selectable atom (i.e., all atoms
in the bodies of the clauses are either $\mathit{true}$ or 
have the form $\underline{a}$), then $E_n$ is called a
\emph{successful explanation} and we refer to
$E_0\rightarrowtail \ldots\rightarrowtail E_n$ 
as a \emph{successful unfolding sequence}.
The \emph{probability of a successful explanation}, $P(E_n)$, 
can be simply obtained as
the product of the probabilities of the probabilistic clauses
in this explanation.

\item If the clauses in $E_n$ contain some selectable atom
which does not unify with the head of any program clause,
then $E_n$ is called a \emph{failing explanation} and it is
discarded from the generation process.
\end{itemize}
By construction, there exists one
successful unfolding sequence associated with each successful
SLD derivation of a query.

\begin{example} \label{ex:smokes_paper}
  Consider again the program in Example~\ref{ex:smokes}, 
  where we now add one  additional ground probabilistic fact:
  \verb$0.1::influences(ann,bob)$. Moreover, assume 
  that predicate $\tt smokes/1$ is visible. Then, we have the following
  successful explanation sequence:
  \[
  \begin{array}{lllll}
    E_0 & = & \{ & \tt  \query(smokes(carl)) ~\mbox{\tt :-}~ smokes(carl) & \}\\
    E_1 & = & \{ & \tt  \query(smokes(carl)) ~\mbox{\tt :-}~ influences(bob,carl),smokes(bob) & \}\\
    E_2 & = & \{ & \tt	\query(smokes(carl)) ~\mbox{\tt :-}~ \underline{influences(bob,carl)},smokes(bob),\\
    & & & \tt  0.3::influences(bob,carl) & \}\\
    E_3 & = & \{ & \tt \query(smokes(carl)) ~\mbox{\tt :-}~ \underline{influences(bob,carl)},\underline{smokes_1(bob)},\\
    & & & \tt  0.3::influences(bob,carl),~~~
                      smokes_1(bob) ~\mbox{\tt :-}~ stress(bob) & \}\\
    E_4 & = & \{ & \tt \query(smokes(carl)) ~\mbox{\tt :-}~ \underline{influences(bob,carl)},\underline{smokes_1(bob)},\\
    & & & \tt  0.3::influences(bob,carl),~~~
                   smokes_1(bob) ~\mbox{\tt :-}~ \underline{stress(bob)},\\
    & & & \tt 0.8::stress(bob) & \}\\
  \end{array}
  \]  
  Therefore, $E_4$ is a successful explanation for the query
  with probability $P(E_4)=\tt 0.24$, and
  can be read as  ``\verb$carl$ smokes because
  \verb$bob$ influences \verb$carl$ (with probability $\tt 0.3$) and
  \verb$bob$ smokes, and \verb$bob$ smokes because he is 
  stressed (with probability $\tt 0.8$)".
  There exists another (less likely) explanation $E'$ as follows:
  \[
  \begin{array}{lllll}
  E' & = & \{ & \tt smokes(carl) ~\mbox{:-}~ \underline{influences(bob,carl)},\underline{smokes_1(bob)}, \\
  & & & \tt smokes_1(bob) ~\mbox{:-}~ \underline{influences(ann,bob)},\underline{smokes_2(ann)},\\
  & & & \tt smokes_2(ann) ~\mbox{:-}~ \underline{stress(ann)}, 
           ~~~0.1::influences(ann,bob),\\
  & & & \tt 0.3::influences(bob,carl),~~
         0.8::stress(ann) & \}
  \end{array}
  \]
  with probability $P(E') = \tt 0.024$. Note that
  the probability of the query in
  $E_4\cup E'$ is the same as in the original program: $\tt 0.2448$
  (and different from $P(E_4)+P(E')$).
\end{example}

\subsubsection{Correctness.}

Our main result states the soundness and completeness of 
successful explanations:\footnote{Proofs of technical results 
can be found in \cite{Vid22tr}.}

\begin{theorem} \label{th:proof}
  Let $\cP$ be a program and $q$ a query. 
  Then, $q$ has a successful SLD 
  derivation in $\cP$ with (ground) computed
  answer $\theta$ iff  there exists a successful explanation $E$
  such that $\query(q)$ has one, and only one, successful SLD
  derivation in $E$
  computing the same answer $\theta$ and using the same 
  probabilistic clauses.
\end{theorem}
As a consequence, we have the following property that states
that the probability of a successful derivation can be obtained
from the product of the probabilistic clauses in the associated
explanation:

\begin{corollary} \label{coro:proof}
  Let $\cP$ be a program and $q$ a query. 
  Then, there is a successful derivation $D$ for $q$ in $\cP$  
  iff there is a successful explanation $E$ with 
  $P(D) = P(E)$. Moreover, $P(E) = P(\query(q))$ in $E$.
\end{corollary}
The above result is an easy consequence of Theorem~\ref{th:proof}
and the fact that $E$ contains all, and only, the probabilistic facts
required for the considered derivation. Note that the success probability
of a query and that of a proof trivially coincide in successful explanations
since only one proof per explanation exists.

Finally, we consider the preservation of the 
success probability of a query in the union of generated explanations:

\begin{theorem} \label{th:query}
  Let $\cP$ be a program and $q$ a query. 
  Let $E_1,\ldots,E_n$ be all an only the successful explanations
  for $q$ in $\cP$. Then, $P(q)$ in $\cP$ is equal to $P(\query(q))$
  in $E_1\cup\ldots\cup E_n$, $n\geq 0$.  
\end{theorem}

\section{The Explanation Generator $\xgen$} \label{sec:xgen}

In order to put into practice the ideas introduced so far, we have
developed a proof-of-concept implementation of the 
explanation generator, called
$\xgen$. The tool has been implemented in SWI Prolog and
includes four modules and approximately one thousand lines of code. The main module implementing the transition rules
of the previous section has some 300 lines of Prolog code.
This module also implements an unfolding \emph{strategy}
that ensures termination in many cases (see the discussion
below). The remaining modules implement some utility predicates
as well as the parser of ProbLog files with visibility  annotations.
The tool can be downloaded from \url{https://github.com/mistupv/xgen}\:.

The tool accepts ProbLog programs containing 
probabilistic facts defined by (not necessarily ground)
facts and rules (i.e., intensional facts). 
The user can also (optionally)
specify which predicates are \emph{visible}
(if any) by means of annotations. 
Furthermore, $\xgen$ accepts duplicated probabilistic facts
as long as the corresponding
predicates are declared \emph{unsafe} using 
an annotation. 
Unsafe atoms are dealt with similarly to
visible atoms when they occur in probabilistic clauses, 
but can be unfolded freely when they appear in a derived clause
(this is why a new annotation is required).
An example specifying an unsafe predicate can be found 
in the above URL.

\begin{figure}[t]
\begin{multicols}{2}
\scriptsize
\begin{verbatim}
$ swipl
Welcome to SWI-Prolog (version 8.2.4)
[...]
?- [xgen].
true.
?- xgen('examples/smokes_paper.pl').
% Explanation #1:
0.3::infl(bob,carl).
0.8::stress(bob).
smokes(bob) :- stress(bob).
smokes(carl) :- infl(bob,carl),smokes(bob).
query(smokes(carl)).
% Success probability: 0.24

% Explanation #2:
0.1::infl(ann,bob).
0.3::infl(bob,carl).
0.8::stress(ann).
smokes(bob) :- infl(ann,bob),smokes0(ann).
smokes(carl) :- infl(bob,carl),smokes(bob).
smokes0(ann) :- stress(ann).
query(smokes(carl)).
% Success probability: 0.024

% No more explanations...

% Combined explanations:
0.1::infl(ann,bob).
0.3::infl(bob,carl).
0.8::stress(ann).
0.8::stress(bob).
smokes(bob) :- stress(bob).
smokes(bob) :- infl(ann,bob),smokes0(ann).
smokes(carl) :- infl(bob,carl),smokes(bob).
smokes0(ann) :- stress(ann).
query(smokes(carl)).
% Success probability: 0.2448

Output files can be found in folder 
"explanations".
\end{verbatim}
\end{multicols}
\caption{A typical session with $\xgen$} \label{fig:xgen}
\end{figure}

As in ProbLog, a query $q$ is added to the program as a fact
of the form $\query(q)$. Let us consider, for instance, the
program from Example~\ref{ex:smokes_paper}, where we now
add the query as the fact 
$\tt \query(smokes(carl))$. If the program is stored
in file \verb$smokes_paper.pl$, a typical session proceeds
as shown in Figure~\ref{fig:xgen}, where the predicate
\verb$influences/2$ is abbreviated to \verb$infl/2$.

In order to deal with cyclic definitions, we have implemented the following
unfolding strategy in $\xgen$: we select
the leftmost atom in the body of a clause that is not 
underlined nor a \emph{variant} 
of any of its (instantiated) \emph{ancestors}.\footnote{Let 
$B_1,a,B_2 \leadsto_\theta (B_1,B,B_2)\theta$ an SLD resolution
step using clause $h\leftarrow B$ and $\mgu(a,h)=\theta$. Then,
$a$ is the direct ancestor of the atoms in $B$. The notion of
ancestor is the transitive closure of this relation.}
For instance, our tool can deal with programs containing cyclic
definitions like 
\begin{verbatim}
  path(X,Y) :- edge(X,Y).
  path(X,Y) :- path(Z,Y), edge(X,Z).
\end{verbatim}
(together with a set of probabilistic facts defining \verb$edge/2$).
Similar strategies have been used, e.g.,
in partial deduction \cite{Bol93}. Although this strategy is clearly
sound---an infinite derivation must necessarily select the same
atom once and again, since the Herbrand universe is 
finite---completeness does not generally hold (a counterexample can be 
found in \cite{BAK91}). As a consequence, our unfolding
strategy could prune some derivations despite the fact that they 
can eventually give rise to a successful SLD derivation.  
Nevertheless, completeness can still be guaranteed for certain
classes of programs, like the \emph{restricted} programs
of \cite{BAK91} that, intuitively speaking, only allow one recursive 
call in the bodies of recursive predicates (as in the definition
of predicate \verb$path/2$ above), so that no infinitely
growing queries can be obtained. The class of restricted
programs is similar to that of \emph{B-stratifiable} logic programs 
\cite{HS04} and its generalization, \emph{strongly regular} logic 
programs \cite{Vid10}, both used in the context of partial 
deduction \cite{LS91}.
As an alternative to using a terminating unfolding strategy as we
do in $\xgen$, one could also consider an implementation of
\emph{tabled} SLD resolution (as in \cite{FBRSGTJR15})
or an iterative deepening strategy (as in \cite{MR17}).
Nevertheless, termination is not decidable for general
logic programs no matter the considered strategy.

\section{Related Work} \label{sec:relwork}

An obvious related work is the definition of 
fold/unfold transformations
for logic programs (see \cite{PP94} and references herein). 
Indeed, rules (\ref{eqn:unfold1}) and (\ref{eqn:unfold3}) 
can be seen as standard unfolding transformations
(except for the differences already mentioned in Section~\ref{sec:unfolding}). 
In general, given a
program $\cP$ and an initial explanation 
$E_0 = \{ \query(q) \leftarrow q\}$, a standard unfolding transformation
on $\cP\cup E_0$ would return 
$\cP\cup E_1\cup\ldots\cup E_n$, where $E_0 \rightarrowtail E_1$,
\ldots, $E_0 \rightarrowtail E_n$
are all the possible unfolding steps 
using rules (\ref{eqn:unfold1}) and (\ref{eqn:unfold3}).
Rules for visible or
probabilistic atoms, i.e., rules (\ref{eqn:unfold2}), (\ref{eqn:unfold4})
and (\ref{eqn:unfold5}), resemble a combination of 
definition introduction and folding, followed by unfolding. 
Nevertheless, a distinctive
feature of our approach is that the computed bindings are shared
by all the clauses in the explanation. 
Indeed, applying
the computed $\mgu$'s to \emph{all} 
clauses in the current explanation 
is sensible in our context since 
all clauses together represent a \emph{single} proof.

To the best of our knowledge, the only previous approach to
defining an unfolding transformation in the context of a probabilistic
logic formalism is that in \cite{Mug00}. However, this work
considers \emph{stochastic logic programs} (SLPs) \cite{Mug96},
a generalization of stochastic grammars and hidden Markov models.
SLPs do not follow the distribution semantics 
(as PLP does). Actually, the unfolding transformation in 
\cite{Mug00} is the standard one for logic programs \cite{PP94}. 
Here, the probability is always preserved by the 
unfolding transformation
because of the way the probability of SLPs is computed 
(i.e., the probability of a query is obtained directly from the 
successful SLD derivations of the query). 

In the context of PLP with a distribution semantics, we are not
aware of any previous work focused on unfolding transformations
or on the generation of explanations other than computing the
MPE \cite{SRVKMJ14} or Viturbi proof \cite{KDRCR11}. 
Actually, we find more similarities between our approach and the
technique called \emph{knowledge-based model construction}
\cite{KdR01} used to compute the grounding of the program
clauses which are relevant for a given query 
(see also \cite{FBRSGTJR15}). However, both the aim and the
technique are different from ours.

Finally, let us mention some recent advances to improve the
quality of explanations in a closely related field: \emph{Answer
Set Programming} (ASP) \cite{BET11}. 
First, \cite{CFM20} presents a tool, $\tt xclingo$, 
for generating explanations from annotated
ASP programs. Annotations are then used
to construct derivation trees containing textual explanations. 
Moreover, the language allows the user to select 
\emph{which} atoms or rules should be included in the explanations.
And, second, \cite{ACCG20} presents 
so-called \emph{justifications} for ASP programs with constraints,
now based on a goal-directed semantics. As in the previous work,
the user can decide the level of detail required in a
justification tree, as well as add annotations to produce
justifications using natural language.
Obviously, our work shares the aim of these papers 
regarding the generation of minimal and understandable
explanations. However, the
considered language and the applied techniques are
different. Nevertheless, we find it very interesting
to extend our work with some of the ideas in 
\cite{ACCG20,CFM20}, e.g., the use of annotations to
produce explanations using natural language.

\section{Concluding Remarks and Future Work} \label{sec:conc}

In this paper, we have presented a novel approach to generate
explanations in the context of PLP
languages like ProbLog \cite{dRKT07}. In particular, and in contrast
to previous approaches, we have proposed explanations to be
represented as programs, one for each proof of a given query. 
In this way,
the user can analyze each (minimal) 
proof separately, understand \emph{why}
the considered prediction (query) is true following the chain
of inferences (an intuitive process) and using a familiar control
structure, that of conditional rules. 
We have formally proved that explanations preserve the 
probability of the original proofs, and that the success probability of
a query can also be computed from the union of the 
generated explanations. A proof-of-concept tool
for generating explanations, $\xgen$, has
been implemented, demonstrating the viability of the approach.

We consider several avenues for future work. On the one hand,
we plan to extend the features of the considered language in order
to include negation,\footnote{Considering negated \emph{ground}
probabilistic facts is straightforward, 
but dealing with negated derived atoms is much more challenging.}
disjunctive probabilistic clauses, evidences, some Prolog built-in's, etc.
This will surely improve the applicability of our approach
and will allow us to carry on an experimental evaluation of
the technique. In particular, we plan to study both the 
scalability of the approach as well as the usefulness of
the generated explanations w.r.t.\ some selected 
case studies.

Another interesting research line consists in allowing the
addition of annotations in program clauses so that natural language 
explanations can be generated (as in \cite{ACCG20,CFM20}).
Finally, we would also like to explore the use of our unfolding
transformation as a pre-processing stage for computing
the probability of a query. 
In particular, when no predicate is declared as visible, our transformation
produces a number of explanations of the form
$\{p_1::a_1,\ldots,~p_n::a_n,~
\query(q) \leftarrow \underline{a_1},\ldots,\underline{a_n}\}$, 
where $p_1::a_1,\ldots,p_n::a_n$ are ground probabilistic facts. 
Apparently, computing the probability of a query from the union of the
generated explanations seems 
much simpler than computing it for an arbitrary program.

\subsubsection*{Acknowledgements.}

I would like to thank the anonymous reviewers 
for their suggestions to improve this paper.

\bibliographystyle{splncs04}

\end{document}